\pdfoutput=1

\documentclass[11pt]{article}
    \makeatletter
\def\@fnsymbol#1{\ensuremath{\ifcase#1\or \dagger\or \ddagger\or
   \mathsection\or \mathparagraph\or \|\or **\or \dagger\dagger
   \or \ddagger\ddagger \else\@ctrerr\fi}}
    \makeatother

\usepackage[preprint]{acl}

\usepackage{times}
\usepackage{latexsym}

\usepackage[T1]{fontenc}

\usepackage[utf8]{inputenc}

\usepackage{microtype}

\usepackage{inconsolata}

\usepackage{graphicx}

\usepackage{graphicx}
\usepackage{booktabs}
\usepackage[utf8]{inputenc} 
\usepackage[T1]{fontenc}    
\usepackage{url}            
\usepackage{booktabs}       
\usepackage{amsfonts}       
\usepackage{nicefrac}       
\usepackage{microtype}      
\usepackage{xcolor}         
\usepackage{multirow}
\usepackage{subcaption}
\usepackage{graphicx}
\usepackage{enumitem}
\usepackage{comment}
\usepackage{amsmath}
\usepackage{amssymb}
\usepackage{algorithm}
\usepackage{algpseudocode}
\usepackage{comment}
\usepackage{CJKutf8}
\usepackage{wrapfig}
\def\eg{\emph{e.g.}} 
\def\ie{\emph{i.e.}}

\newcommand{\sub}[1]{\textcolor{red}{#1}}
\newcommand{\del}[1]{\textcolor{blue}{#1}}
\newcommand{\ins}[1]{\textcolor{brown}{#1}}
\DeclareMathOperator*{\argmax}{arg\,max}

\title{Towards Online Continuous Sign Language Recognition and Translation}

\author{
 \textbf{Ronglai Zuo\textsuperscript{1}}\qquad
 \textbf{Fangyun Wei\textsuperscript{2}\thanks{Corresponding author.}}\qquad
 \textbf{Brian Mak\textsuperscript{1}}
\\
 \textsuperscript{1}The Hong Kong University of Science and Technology\quad
 \textsuperscript{2}Microsoft Research Asia
\\
\texttt{
   \href{mailto:rzuo@connect.ust.hk}{\textcolor{black}{rzuo@connect.ust.hk}} \quad
   \href{mailto:fawe@microsoft.com}{\textcolor{black}{fawe@microsoft.com}} \quad
   \href{mailto:mak@cse.ust.hk}{\textcolor{black}{mak@cse.ust.hk}}
 }
}

\begin{document}
\maketitle

\begin{abstract}
Research on continuous sign language recognition (CSLR) is essential to bridge the communication gap between deaf and hearing individuals. Numerous previous studies have trained their models using the connectionist temporal classification (CTC) loss. During inference, these CTC-based models generally require the entire sign video as input to make predictions, a process known as offline recognition, which suffers from high latency and substantial memory usage. In this work, we take the first step towards online CSLR. Our approach consists of three phases: 1) developing a sign dictionary; 2) training an isolated sign language recognition model on the dictionary; and 3) employing a sliding window approach on the input sign sequence, feeding each sign clip to the optimized model for online recognition. Additionally, our online recognition model can be extended to support online translation by integrating a gloss-to-text network and can enhance the performance of any offline model. With these extensions, our online approach achieves new state-of-the-art performance on three popular benchmarks across various task settings. Code and models are available at \href{https://github.com/FangyunWei/SLRT}{https://github.com/FangyunWei/SLRT}.

\end{abstract}
\vspace{-3mm}
\section{Introduction}
\vspace{-1mm}
\begin{figure*}[t]
\begin{subfigure}{1.0\textwidth}
\centering
\includegraphics[width=\textwidth]{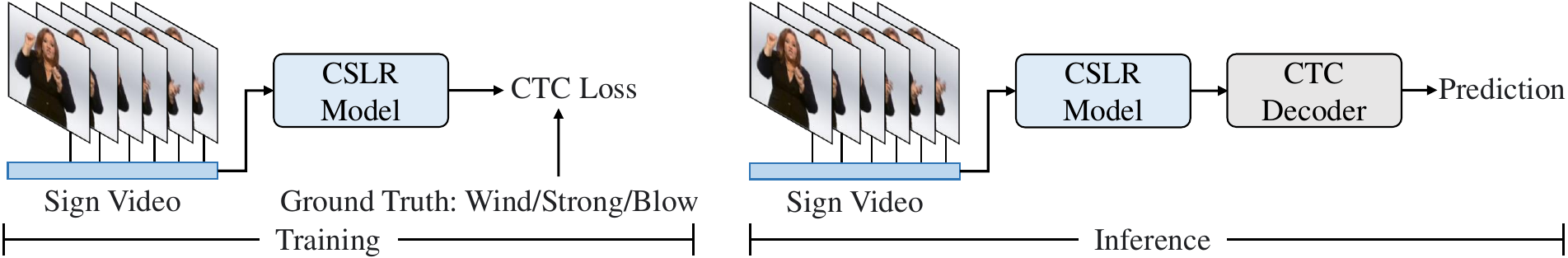}
\caption{Training and inference of previous \textit{offline} recognition models that are trained using the CTC loss. These models require access to the \textit{entire} sign video before they can make predictions.}
\label{fig:teaser_ctc}
\end{subfigure}

\begin{subfigure}{1.0\textwidth}
\centering
\includegraphics[width=\textwidth]{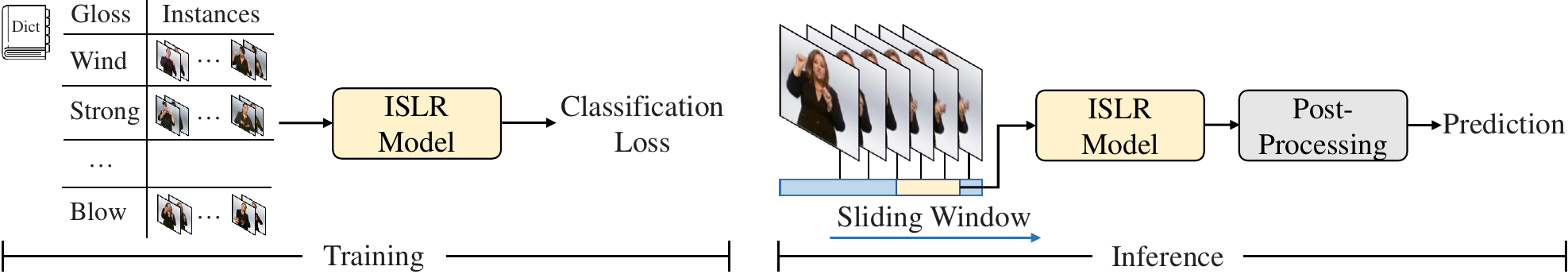}
\caption{Training and inference in our \textit{online} approach. We utilize a pre-trained CSLR model to segment all continuous sign videos into isolated sign clips. This process creates a dictionary for each CSLR dataset, which supports the subsequent training of an ISLR model. During inference, we apply a sliding window to the input sign stream and perform on-the-fly predictions. The function of post-processing is to eliminate duplicates and background predictions.}
\label{fig:teaser_ours}
\end{subfigure}
\vspace{-5mm}
\caption{Illustration of (a) the offline recognition scheme and (b) the proposed online framework.
}
\label{fig:teaser}
\vspace{-3mm}
\end{figure*}

Sign languages are visual languages conveyed through hand shapes, body movements, and facial expressions. The domain of sign language recognition (SLR) \cite{Jiao_2023_ICCV, chentwo} has recently attracted considerable attention, particularly for its potential to bridge the communication gap between the hearing and deaf communities. SLR can be categorized into isolated sign language recognition (ISLR)~\cite{hu2023signbert+, zuo2023natural} and continuous sign language recognition (CSLR)~\cite{chentwo, zheng2023cvt}. ISLR, a supervised classification task, aims to accurately predict the gloss\footnote{A gloss is a label associated with an isolated sign.} for each individual sign. In contrast, as no annotations of sign boundaries are provided, CSLR is a weakly supervised task. In this context, a well-optimized model is able to predict a sequence of glosses from a continuous sign video containing multiple signs. 
Compared to ISLR, CSLR is more challenging but also more practical. The primary objective of this work is to develop an online CSLR system.

Drawing inspiration from the advancements in speech recognition~\cite{deep_speech2}, numerous CSLR models are trained using the established connectionist temporal classification (CTC) loss~\cite{ctc} with sentence-level annotations. During inference, these models typically process the \textit{entire} sign video to make predictions, leading to issues like high latency and significant memory consumption. This method is known as offline recognition, as depicted in Fig.~\ref{fig:teaser_ctc}. Unlike modern speech recognition systems, which can recognize spoken words on the fly, CSLR still lags behind due to the lack of practical online recognition solutions, which are essential in real-world scenarios such as live conversations or emergency situations. Although CTC-based methods can be adapted for online recognition using a sliding window technique, our empirical findings show that the discrepancy between training (using entire, untrimmed sign videos) and inference (using short, trimmed sign clips) results in suboptimal performance.

In this paper, we take the first step towards practical online CSLR. Instead of directly training with the CTC loss on a CSLR dataset, we optimize an ISLR model using classification losses on a dictionary containing all glosses from the target CSLR dataset. During inference, we apply a sliding window to a given sign video stream and process each video clip through the well-optimized ISLR model to obtain corresponding predictions. This approach aligns training and inference by utilizing short video clips as input for both. However, using a sliding window with a small stride may lead to multiple scans of the same sign, resulting in repetitive predictions. To mitigate this, we introduce an effective post-processing technique to eliminate duplicate predictions. We also consider co-articulations, which are the transitional movements of the body and hands between consecutive signs in a continuous video. Since these movements are generally meaningless, we assign them to an additional background category, and predictions in this category are discarded during post-processing. Our online method is illustrated in Fig.~\ref{fig:teaser_ours}.

In our methodology, we train the ISLR model using a sign dictionary. Existing CSLR datasets, such as Phoenix-2014~\cite{2014}, Phoenix-2014T~\cite{2014T}, and CSL-Daily~\cite{zhou2021improving}, lack such dictionaries. However, a pre-trained CSLR model utilizing CTC loss can effectively segment continuous sign videos into individual isolated signs~\cite{dnf,Wei_2023_ICCV,zuo2024simple}. This process, known as CTC forced alignment, is a well-established technique in the speech community for accurately aligning transcripts to speech signals~\cite{ctc}. Therefore, we use the state-of-the-art CSLR model, TwoStream-SLR~\cite{chentwo}, as the sign segmentor for any CSLR dataset. This approach allows us to create a sign dictionary that aligns with the glossary of the respective CSLR dataset. During inference, a fixed-length sliding window may inadvertently include both a sign and its co-articulations. To better align the training and the inference, we generate a set of augmented signs from each isolated sign by trimming video segments surrounding it. This procedure also significantly enriches the training data.

Different signs typically exhibit various durations. 
Personal habits of signers, \eg, signing speed, can also amplify this issue. 
This necessitates the model's adaptability to variations in sign durations, especially when a sliding window includes both a sign and co-articulations. We introduce a saliency loss, which compels the model to focus predominantly on the foreground signs while minimizing the influence of the co-articulations. The implementation is simple—we adopt an auxiliary classification loss on the pooled feature of the foreground parts.

While our method is primarily designed for online CSLR, it also shows promise for online sign language translation (SLT) and enhancing offline CSLR models. We start by implementing an additional gloss-to-text network, applying the wait-$k$ policy~\cite{ma2019stacl} tailored for simultaneous (online) machine translation. This allows for online SLT by gradually feeding the gloss predictions generated by our online CSLR model into the wait-$k$ gloss-to-text network. Furthermore, our online CSLR model can facilitate offline CSLR models in performance. This is achieved by incorporating a lightweight adapter into our frozen online model and combining the adapter-generated features with those extracted by a pre-trained offline CSLR model.

Our contributions can be summarized as follows:

\vspace{-8pt}
\begin{itemize}
\setlength{\itemsep}{0pt}
\setlength{\parsep}{0pt}
\setlength{\parskip}{1pt}
    \item \textbf{One framework}. We introduce an innovative online CSLR framework that slides an ISLR model over a sign video stream. To enhance the ISLR model training, we further propose several techniques such as sign augmentation, gloss-level training, and saliency loss. 
    \item \textbf{Two extensions}. First, we implement online SLT by integrating a wait-$k$ gloss-to-text network. Second, we extend the online CSLR framework to boost the performance of offline CSLR models through a lightweight adapter.
    \item \textbf{Performance.} Our online approach along with the two extensions establishes new state-of-the-art results on three widely adopted benchmarks: Phoenix-2014, Phoenix-2014T, and CSL-Daily, under various task settings.
\end{itemize}

\vspace{-3mm}
\section{Related Works}
\vspace{-1mm}

\noindent\textbf{CSLR, ISLR, and SLT.} 
Since only sentence-level annotations are provided for CSLR, most CSLR works \cite{c2slr,zheng2023cvt,vac,chentwo,hu2023continuous,niu2024hong} adopt the well-established CTC loss, which is proven effective in speech recognition \cite{deep_speech2}, to train their models. These CTC-based models have achieved satisfactory offline CSLR performance. However, there is a notable performance drop in online scenarios due to the discrepancy between training on long, untrimmed videos and inference on short clips. To address this, FCN \cite{fcn} proposes a fully convolutional network with a small receptive field for preliminary online CSLR efforts. However, FCN is still trained on long videos, maintaining the training-inference discrepancy, and its performance remains suboptimal. In this work, we propose a novel approach by training an ISLR model on a sign dictionary, enabling effective online inference through a sliding window strategy.

ISLR is a classification task and has been explored in numerous recent works \cite{ hu2023signbert+, zuo2023natural, Lee_2023_ICCV}. Some CSLR models \cite{dnf, iopt, stmc} adopt the idea of ISLR to iteratively train their feature extractors, a process also known as stage optimization \cite{dnf}. In this work, our ISLR model not only achieves promising results in online recognition but also boosts the offline models using a lightweight adapter network.

Taking a step further, SLT \cite{chen2024factorized,yu2023efficient,gan2023contrastive,lin2023gloss,zhang2023sltunet} involves translating sign languages into spoken languages. This task is commonly approached as an NMT problem, employing a visual encoder followed by a seq2seq translation network. Similar to CSLR, online SLT remains largely unexplored.

\noindent\textbf{Sign Spotting.} 
Given an isolated sign, the goal of sign spotting is to identify whether and where it has been signed in a continuous sign video \cite{varol2022scaling}.
Modern sign spotting works typically rely on extra cues, including external dictionaries \cite{momeni2020watch}, mouthings \cite{albanie2020bsl}, or Transformer attention \cite{varol2021read}. However, these cues can be either difficult to obtain (\eg, dictionaries) or unreliable (\eg, mouthings). Sign spotting is typically used to enrich the training source for ISLR and few works validate the spotting task in the context of CSLR.

\noindent\textbf{Online Speech Recognition.}
Practical online speech recognition systems have been studied in numerous works \cite{pratap2020scaling, he2019streaming, an22_interspeech}. In these studies, model architectures vary, including CNN \cite{pratap2020scaling}, Transformer \cite{miao2020transformer}, and a combination of them \cite{an22_interspeech}. Additionally, multiple optimization frameworks are explored, such as CTC \cite{pratap2020scaling}, seq2seq \cite{fan2019online}, or a hybrid of these \cite{miao2019online}. 
Unlike online speech recognition, online CSLR remains under-explored. This work takes the first step towards building a practical online CSLR framework.

\vspace{-2mm}
\section{Method}
\vspace{-1mm}

\begin{figure*}[t]
\begin{subfigure}{1.0\textwidth}
\centering
\includegraphics[width=\textwidth]{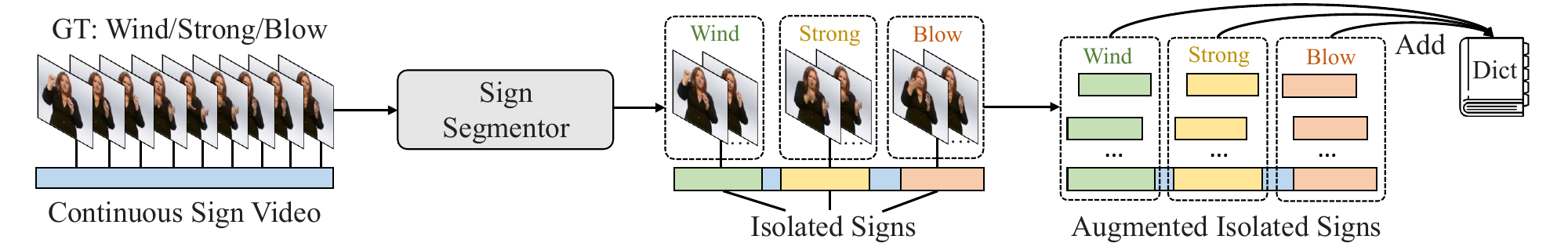}
\caption{Dictionary construction. We adopt a pre-trained CSLR model as the sign segmentor to segment each continuous sign video into its constituent isolated signs, referred to as pseudo ground truths. Subsequently, we create augmented signs by cropping clips around each pseudo ground truth. Both segmented isolated signs (pseudo ground truths) and augmented signs are then incorporated into our dictionary.}
\label{fig:overview_a}
\end{subfigure}

\begin{subfigure}{1.0\textwidth}
\centering
\includegraphics[width=\textwidth]{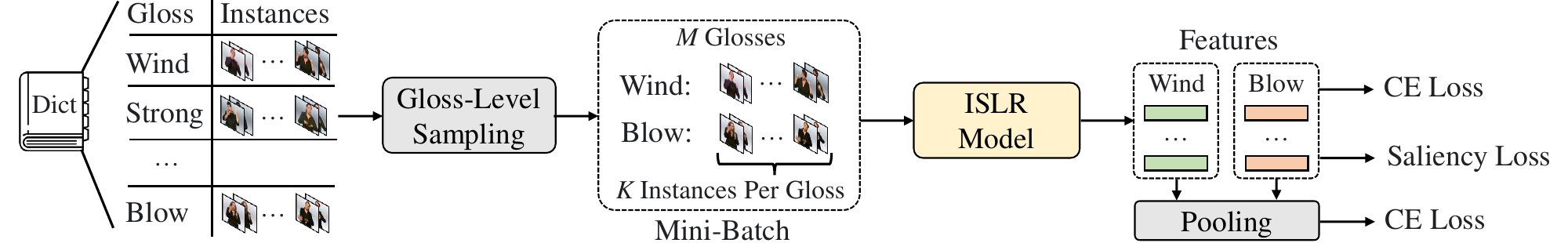}
\caption{Training of ISLR model. Our model is trained on the created dictionary under the supervision of the cross-entropy (CE) loss and the proposed saliency loss (Fig.~\ref{fig:overview_c}) at both instance and gloss levels.}
\label{fig:overview_b}
\end{subfigure}

\begin{subfigure}{1.0\textwidth}
\centering
\includegraphics[width=\textwidth]{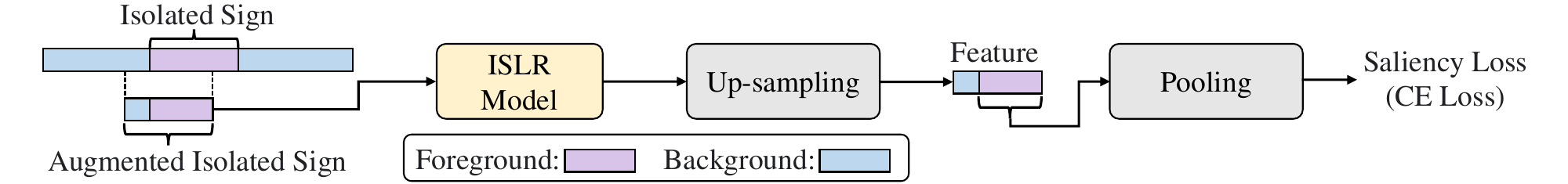}
\caption{Saliency loss. It drives the model to focus more on the foreground sign instead of meaningless co-articulations (background).}
\label{fig:overview_c}
\end{subfigure}

\begin{subfigure}{1.0\textwidth}
\centering
\includegraphics[width=\textwidth]{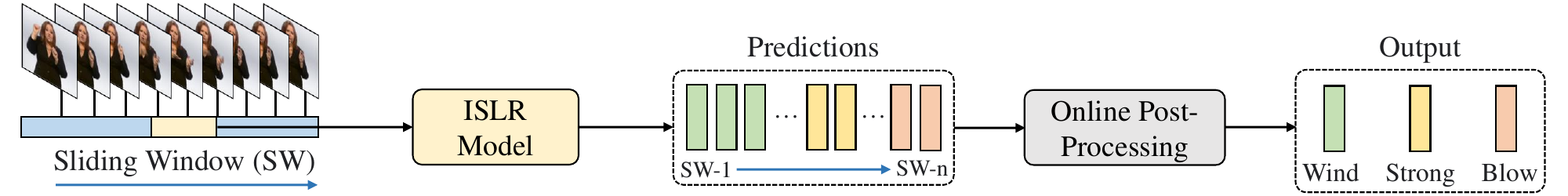}
\caption{Online inference. To enable online recognition, we utilize a sliding window approach on the input sign stream. The objective of our post-processing is to eliminate duplicate predictions and background (blank) predictions.}
\label{fig:overview_d}
\end{subfigure}

\vspace{-2mm}
\caption{Overview of our methodology.}
\label{fig:overview}
\vspace{-4mm}
\end{figure*}

An overview of our online framework is shown in Fig. \ref{fig:overview}. We first build a sign dictionary with the aid of a sign segmentor, \ie, a pre-trained CSLR model (Sec. \ref{sec:dict}). Then, we train an ISLR model on this dictionary, employing dedicated loss functions at both the instance and gloss levels (Sec. \ref{sec:train}). This is followed by a demonstration of online inference using the optimized ISLR model (Sec. \ref{sec:inference}). Finally, we present two extensions (Sec. \ref{sec:ext}): (1) enabling online SLT with a wait-$k$ gloss-to-text network; (2) boosting the performance of an offline model using our online model.

\vspace{-2mm}
\subsection{Dictionary Construction}
\vspace{-1mm}
\label{sec:dict}
\textbf{Sign Segmentor.}
Existing CSLR datasets \cite{2014, 2014T, zhou2021improving} only provide sentence-level gloss annotations, lacking labels for the temporal boundaries of each isolated sign. Inspired by the observation that a well-trained CTC-based CSLR model can identify the approximate boundaries of the isolated signs in a continuous sign video—by searching the most probable alignment path with respect to the ground truth (GT)~\cite{dnf}—we adopt the state-of-the-art CSLR model, TwoStream-SLR~\cite{chentwo}, as the sign segmentor. This model segments each continuous sign video into a sequence of isolated signs, as depicted in Fig. \ref{fig:overview_a}.

We collect these segmented signs (pseudo GT) to form a dictionary $\mathcal{D}$. Each sign $s \in \mathcal{D}$ is expressed as a quadruple $(\boldsymbol{V}, t_b, t_e, g)$, where $\boldsymbol{V}$ is the corresponding continuous sign video, $t_b$ and $t_e$ denote the beginning and ending frame indexes of sign $s$ in $\boldsymbol{V}$, and $g$ is the gloss label. Note that $g\in \mathcal{V}_g \cup \{\varnothing\}$, where $\mathcal{V}_g$ is the gloss vocabulary and $\varnothing$ is the blank (background) class. The segmentation process is detailed in Sec. \ref{sec:sign_seg}.

\noindent\textbf{Sign Augmentation.}
During inference, a sliding window may inadvertently include both a sign and its co-articulations. To better align the training and the inference, we generate a set of augmented signs for each pseudo GT. This is done by cropping clips around each $s \in \mathcal{D}$, as shown in Fig.~\ref{fig:overview_a}. For each pseudo GT sign $s=(\boldsymbol{V}, t_b, t_e, g)$ appearing in a sign video $\boldsymbol{V}$, we generate $t_e - t_b + 1$ augmented instances $\{(\boldsymbol{V}, i-W/2, i+W/2-1, g)\}_{i=t_b}^{t_e}$ around $s$, where $W$ ($W=16$ by default) is the window size. These yielded instances are then added to the dictionary, thereby significantly enhancing the training source. Consequently, the final sign dictionary consists of $N_g$ glosses, each linked to a set of sign instances that include both pseudo GT signs $\{s\}$ and augmented signs $\{\hat{s}\}$.

\vspace{-1mm}
\subsection{ISLR Model}
\vspace{-1mm}
\label{sec:train}
This section delineates the training methodology and the associated loss functions utilized for the ISLR model. Following TwoStream-SLR \cite{chentwo}, the backbone comprises two parallel S3D~\cite{xie2018rethinking} networks, which model RGB sign videos and human keypoints, respectively. The input sign video spans $W$ frames.

\noindent\textbf{Mini-Batch Formation.}
In the traditional classification task, instances from a training set are randomly selected to form a mini-batch. This sampling strategy is referred to as instance-level sampling. In this work, we empirically discover that the gloss-level sampling (our default strategy) yields better performance. As illustrated in Fig. \ref{fig:overview_b}, we initially sample $M$ glosses from the dictionary. For each gloss, we then sample $K$ instances to form a mini-batch, resulting in an effective batch size of $M\times K$. In our implementation, the $K$ instances sampled for each gloss can be either a pseudo GT sign or its augmentations as described in Sec.~\ref{sec:dict}. Our technique shares a similar spirit with batch augmentation (BA) \cite{hoffer2020augment}, which augments a mini-batch multiple times. Our gloss-level sampling differs by employing ``temporally jittered'' instances around the pseudo GT signs to form a training batch, instead of directly augmenting the pseudo GT as in BA. Nevertheless, our sampling strategy still retains the benefits of BA, such as decreased variance reduction.

\noindent\textbf{Loss Functions.}
Given a mini-batch with a size of $M\times K$, let $p_j^i$ denote the posterior probability of the sample with gloss index $i\in [1,M]$ and instance index $j\in [1,K]$. The classification loss of our ISLR model is composed of two parts: 1) an instance-level cross-entropy loss ($\mathcal{L}^I_{ce}$) applied across $M\times K$ instances; 2) a gloss-level cross-entropy loss ($\mathcal{L}^G_{ce}$) applied over $M$ glosses to learn more separable representations. The two losses can be formulated as:
\vspace{-2mm}
\begin{equation}
\begin{split}
    &\mathcal{L}^I_{ce} = -\frac{1}{M\times K} \sum_{i=1}^M \sum_{j=1}^K \log p_j^i, \\ &\mathcal{L}^G_{ce} = -\frac{1}{M} \sum_{i=1}^M \log \frac{1}{K} \sum_{j=1}^K p_j^i.
\end{split}
\end{equation}
\vspace{-2mm}

\noindent\textbf{Saliency Loss.}
Our ISLR model processes sign clips with a fixed length, but the foreground regions in these clips can vary. To address this, we devise a saliency loss that encourages the model to prioritize the foreground sign and disregard the background signs (co-articulations). An illustration of the proposed saliency loss is shown in Fig. \ref{fig:overview_c}. In detail, for a training sample $\hat{s}=(\boldsymbol{V}, \hat{t}_b, \hat{t}_e, g)$, which is an augmented instance of pseudo GT $s=(\boldsymbol{V}, t_b, t_e, g)$, we input it into our ISLR model. This process yields its encoded feature $\boldsymbol{f} \in \mathbb{R}^{T_s/\alpha\times C}$, where $T_s = \hat{t}_e - \hat{t}_b + 1$ is the clip length, $\alpha=8$ is the down-sampling factor of the neural network, and $C$ denotes the channel dimension. Next, we up-sample $\boldsymbol{f}$ to $\boldsymbol{f}_u \in \mathbb{R}^{\beta T_s/\alpha \times C}$ using an up-sampling factor $\beta$ ($\beta=4$ by default). The overall scaling factor thus becomes $\beta/\alpha$. Without loss of generality, assuming that $\hat{t}_b\leq t_b\leq \hat{t}_e \leq t_e$, the foreground area starts from the $t_b$-th frame and ends at the $\hat{t}_e$-th frame. We then can generate the foreground feature $\boldsymbol{f}_f \in \mathbb{R}^{C}$ by pooling $\boldsymbol{f}_u[\lceil\beta t_b/\alpha\rceil:\lfloor\beta \hat{t}_e/\alpha\rfloor,~:]$ along the temporal dimension. 

Finally, the saliency loss $\mathcal{L}_{s}$ is implemented as a cross-entropy loss over the probability yielded from $\boldsymbol{f}_{f}$. Similar to $\mathcal{L}_{ce}^I$ and $\mathcal{L}_{ce}^G$, our saliency loss is imposed at both instance and gloss levels, denoted as $\mathcal{L}_{s}^I$ and $\mathcal{L}_{s}^G$, respectively.

\noindent\textbf{Overall Loss Function.} It is implemented as the summation of the classification loss and the saliency loss at both instance and gloss levels: $\mathcal{L} = \mathcal{L}_{ce}^I + \mathcal{L}_{ce}^G + \mathcal{L}_{s}^I + \mathcal{L}_{s}^G$.

\subsection{Online Inference}
\label{sec:inference}

As shown in Fig. \ref{fig:overview_d}, the online inference is implemented using a sliding-window strategy with a stride of $S$. Generally, sliding-window approaches produce duplicate predictions, as they may scan the same sign multiple times. Therefore, post-processing is always necessary. The pseudo code of our online post-processing is provided in Alg. \ref{alg:post} in the appendix. The algorithm has two key functions: (1) voting-based deduplication (Line \ref{code:voting}), and (2) background elimination (Line \ref{code:drop_blk}). 
Please refer to Sec. \ref{sec:inf_detail} for more details.

\subsection{Extensions}
\label{sec:ext}

\begin{figure}[t]
\centering
\includegraphics[width=\linewidth]{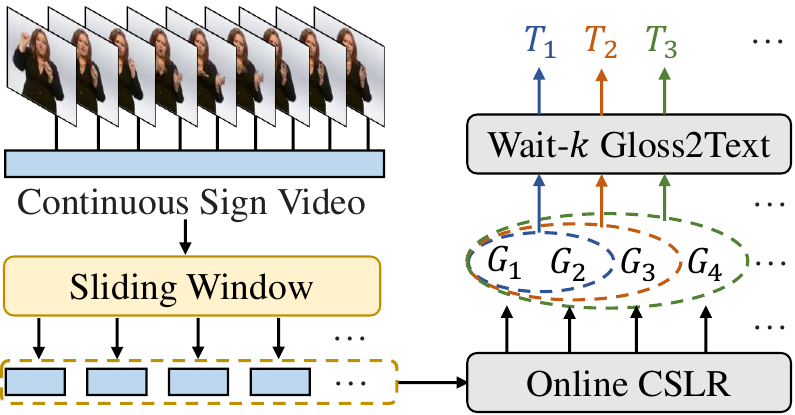}
\vspace{-7mm}
\captionof{figure}{Appending a gloss-to-text network with the wait-$k$ policy onto our online CSLR model enables online SLT. Circles and arrows distinguished by varied colors indicate translation outcomes at distinct timings.}
\label{fig:slt}
\vspace{-3mm}
\end{figure}

\noindent\textbf{Online Sign Language Translation.} 
As shown in Fig.~\ref{fig:slt}, we append an additional gloss-to-text network~\cite{chentwo} with the wait-$k$ policy \cite{ma2019stacl} onto our online CSLR model to enable online SLT. This wait-$k$ policy enables text predictions after seeing $k$ glosses ($k=2$ following~\cite{yin2021simulslt}). During the inference phase, gloss predictions produced by our online CSLR model are sequentially fed into the well-optimized gloss-to-text network, to produce translation results.

\noindent\textbf{Promote Offline Models with Online Model.}
Our online CSLR model can also enhance the performance of offline models. As shown in Fig. \ref{fig:adapter}, consider two well-optimized CSLR models: our online model and an existing offline model. Let $\boldsymbol{\hat{f}}$ and $\boldsymbol{\tilde{f}}$ denote the features extracted by the online model and the offline model, respectively. To first align the dimensions of the two features, we attach a lightweight adapter network—comprising a down-sampling layer and a 2-layer MLP—to the online model. This network projects $\boldsymbol{\hat{f}}$ to $\boldsymbol{\bar{f}}$, matching the dimension of $\boldsymbol{\tilde{f}}$. We then fuse $\boldsymbol{\bar{f}}$ and $\boldsymbol{\tilde{f}}$ using a weighted sum operation: $\boldsymbol{f}_{fuse} = \lambda\cdot\boldsymbol{\bar{f}} + (1-\lambda)\cdot\boldsymbol{\tilde{f}}$, where $\lambda$ is a trade-off hyper-parameter set to 0.5 by default. Finally, $\boldsymbol{f}_{fuse}$ is fed into a classification head supervised by the CTC loss. The training is extremely efficient since the parameters of both online and offline models are frozen. We adopt TwoStream-SLR~\cite{chentwo} as the offline model due to its exceptional performance.

\vspace{-2mm}
\section{Experiments}
\vspace{-1mm}
\subsection{Implementation Details}
\vspace{-1mm}
\textbf{Datasets.}
We evaluate our method on three widely-adopted datasets: Phoenix-2014 (P-2014) \cite{2014}, Phoenix-2014T (P-2014T) \cite{2014T}, and CSL-Daily (CSL) \cite{zhou2021improving}. \textit{P-2014} is a German sign language dataset consisting of 5,672/540/629 samples in the training, development (dev), and test set, respectively, with a vocabulary size of 1,081 glosses. \textit{P-2014T} is an extension of P-2014, which consists of 1,066 glosses and 7,096/519/642 samples in the training, dev, and test set. \textit{CSL} is the latest Chinese sign language dataset with a vocabulary size of 2,000 glosses. There are 18,401/1,077/1,176 samples in its training, dev, and test set.

\noindent\textbf{Evaluation Metrics.}
Following \cite{chentwo}, we use word error rate (WER), which measures the dissimilarity between the prediction and the GT, as the evaluation metric for CSLR. A lower WER indicates better performance. 
For SLT, we report BLEU-4 scores computed by SacreBLEU (v1.4.2) \cite{sacrebleu}.

\noindent\textbf{Training.}
Our ISLR model is trained with an effective batch size of $4\times6$ (4 glosses and 6 instances per gloss), for 100 epochs. 
We use a cosine annealing schedule and an Adam optimizer \cite{adam} with a weight decay of $1e^{-3}$ and an initial learning rate of $6e^{-4}$. 
When fine-tuning the adapter network and classification head, we use a smaller learning rate of $1e^{-4}$ and fewer epochs of 40. We set $\lambda=0.5$.

\noindent\textbf{Inference.}
Online inference is implemented using a sliding window approach. We set $W=16, S=1, B=7$ in Alg. \ref{alg:post}. For both offline inference and CTC-based online inference, a CTC decoder with a beam width of 5 is used following~\cite{chentwo}. More details and studies on hyper-parameters are available in the appendix.

\begin{figure}[t]
\centering
\includegraphics[width=\linewidth]{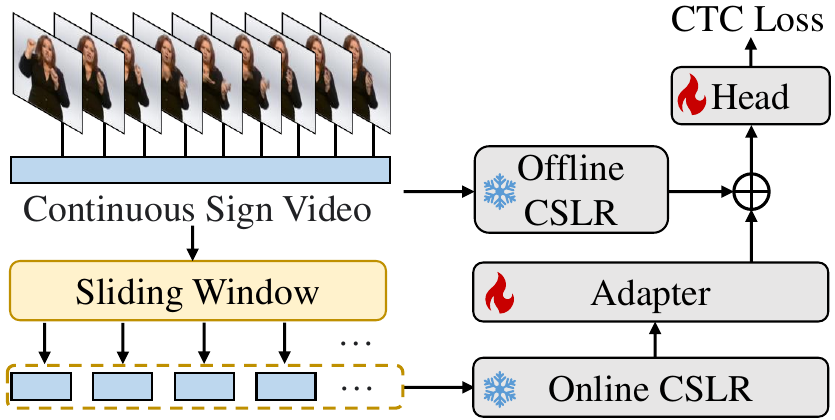}
\vspace{-7mm}
\captionof{figure}{Boosting an offline model with our online model. A lightweight adapter fuses the features of two well-trained CSLR models, one offline and one online. The parameters of both CSLR models remain frozen.}
\label{fig:adapter}
\vspace{-3mm}
\end{figure}

\vspace{-1mm}
\subsection{Comparison with SOTA Methods}
\vspace{-1mm}
\begin{table*}[t]
\small
\setlength\tabcolsep{5pt}
    \centering
    \begin{tabular}{l|c|cc|cc|cc}
    \toprule
    \multirow{2}{*}{Method} & \multirow{2}{*}{Window Size} & \multicolumn{2}{c|}{P-2014} & \multicolumn{2}{c|}{P-2014T} & \multicolumn{2}{c}{CSL}\\
    & & Dev$\downarrow$ & Test$\downarrow$ & Dev$\downarrow$ & Test$\downarrow$ & Dev$\downarrow$ & Test$\downarrow$ \\

    \midrule
    \multirow{4}{*}{FCN$^*$ \cite{fcn}} & 40 & 29.2 & 28.9 & 29.0 & 29.6 & 44.7 & 44.7 \\
    & 32 & 30.5 & 30.2 & 30.4 & 30.8 & 48.9 & 50.2 \\
    & 24 & 32.5 & 32.8 & 32.9 & 34.7 & 55.6 & 56.3 \\
    & 16 & 36.5 & 36.4 & 38.8 & 39.1 & 72.3 & 72.8 \\
    
    \midrule
    \multirow{4}{*}{TwoStream-SLR \cite{chentwo}} & 40 & 23.6 & 23.7 & 23.1 & 23.9 & 43.0 & 43.7 \\
    & 32 & 25.1 & 25.0 & 24.7 & 26.0 & 52.7 & 53.7 \\
    & 24 & 26.8 & 26.6 & 28.8 & 29.6 & 68.4 & 69.2 \\
    & 16 & 30.3 & 31.6 & 38.4 & 39.3 & 101.4 & 103.3 \\

    \midrule
    Ours & 16 & \textbf{22.6} & \textbf{22.1} & \textbf{22.2} & \textbf{22.1} & \textbf{30.2} & \textbf{29.3} \\

    \bottomrule
    \end{tabular}
    \vspace{-2mm}
    \caption{Comparison with other \textit{online} CSLR methods across three benchmarks. With the aid of a sliding window, TwoStream-SLR \cite{chentwo} (state-of-the-art offline model) is capable to fulfill online recognition. $^*$: Due to the unavailability of the source code, we reimplement FCN \cite{fcn}, a preliminary attempt for online CSLR. We report their performance using the WER\% metric.}
    \label{tab:online}
    \vspace{-3mm}
\end{table*}

\noindent\textbf{Online Recognition.}
Almost all previous CSLR works are trained under the supervision of the CTC loss~\cite{ctc}. During inference, these approaches generally process an \textit{entire} sign video to generate predictions, \ie, offline recognition. These CTC-based approaches can be simply adapted to online recognition by employing a sliding window strategy to the input sign stream. To decode the predictions of the current window $U$, the initial step involves feeding $U$ into the CSLR model, which yields a probability map $\boldsymbol{P}$. The prediction of $U$ is obtained from the CTC decoder, which considers $\boldsymbol{P}$ and the last decoding state of the preceding window. Refer to the original implementation~\cite{ctc,online_ctc} for more details. We implement the online inference for the previously best-performing offline model, TwoStream-SLR, equipped with the same post-processing algorithm. The comparison with the online TwoStream-SLR is shown in Tab. \ref{tab:online}. The performance of online TwoStream-SLR significantly degrades compared to its offline counterpart. We hypothesize that this decline in performance is due to the discrepancy between training (on untrimmed sign videos) and inference (on short sign clips). Even using a larger window size of 40, the performance gap remains over 5\% on P-2014 and 18\% on CSL. This gap is particularly pronounced on CSL, which we attribute to the longer duration of test videos in CSL. In contrast, our method directly optimizes an ISLR model and feeds each sliding window into this well-optimized model during inference, thereby aligning training and inference processes. Our approach outperforms online TwoStream-SLR with a window size of 16 by 9.5/17.2/74.0\% across the three datasets.

\begin{table}[t]
\small
    \centering
    \begin{tabular}{l|c|cc|cc}
    \toprule
        \multirow{2}{*}{Method} & \multirow{2}{*}{\shortstack{Win.\\Size}} & \multicolumn{2}{c|}{P-2014T} & \multicolumn{2}{c}{CSL} \\
        & & Dev$\uparrow$ & Test$\uparrow$ & Dev$\uparrow$ & Test$\uparrow$ \\
    \midrule
        SimulSLT & N/A & 22.85 & 23.14 & -- & 13.88$^*$ \\
    \midrule
        \multirow{4}{*}{TwoStream} & 40 & 22.80 & 22.64 & 18.54 & 17.98 \\
        & 32 & 22.23 & 22.01 & 16.32 & 16.23 \\
        & 24 & 22.19 & 19.92 & 13.66 & 13.49 \\
        & 16 & 18.36 & 18.81 & 10.40 & 9.98 \\
    \midrule
        Ours & 16 & \textbf{23.75} & \textbf{23.69} & \textbf{21.20} & \textbf{20.63} \\
    \bottomrule
    \end{tabular}
    \vspace{-2mm}
    \caption{Comparison with other \textit{online} SLT methods on two benchmarks. For fair comparison, we use the same wait-$k$ gloss-to-text network for both TwoStream Network \cite{chentwo} and our method. $^*$ denotes reimplementation results in \cite{sun2024adaptive}.}
    \label{tab:sota_slt}
    \vspace{-3mm}
\end{table}

FCN \cite{fcn} presents a preliminary attempt for online CSLR, using a fully convolutional network with a small receptive field. However, its evaluation lacks real-world applicability. The authors simulate the online scenario by either concatenating multiple sign videos or splitting a single video into a predefined number of chunks. To ensure a fair comparison under a realistic scenario, we reimplement FCN and achieve offline recognition WERs of 23.9/24.2\% and 23.0/24.5\% on P-2014 and P-2014T, respectively. These results are comparable to those reported in the original FCN paper. When evaluated in the online context, our model consistently outperforms FCN across all three benchmarks, as shown in Tab. \ref{tab:online}.

\noindent\textbf{Online Translation.}
The pioneering effort in online SLT is made by SimulSLT \cite{yin2021simulslt}. Our approach diverges from SimulSLT in three main aspects: 1) instead of using a masked Transformer like SimulSLT to encode sign videos, we incorporate an ISLR model for encoding sign clips; 2) for inference, where SimulSLT relies on a boundary predictor to generate word boundaries, we adopt a more straightforward sliding window strategy; 3) unlike SimulSLT, which is tailored exclusively for online SLT, our model is versatile enough to accommodate both online CSLR and SLT. As shown in Tab. \ref{tab:sota_slt}, by integrating the wait-$k$ gloss-to-text network into our online CSLR model, we observe superior BLEU-4 scores in comparison to SimulSLT. Furthermore, our translation model also outperforms the online TwoStream Network, despite using the same gloss-to-text network.

\begin{table}[t]
\small
\setlength\tabcolsep{1pt}
    \centering
    \resizebox{\linewidth}{!}{
    \begin{tabular}{l|cc|cc|cc}
    \toprule
        \multirow{2}{*}{Method}  & \multicolumn{2}{c|}{P-2014} & \multicolumn{2}{c|}{P-2014T} & \multicolumn{2}{c}{CSL}  \\
        & Dev$\downarrow$ & Test$\downarrow$ &  Dev$\downarrow$ & Test$\downarrow$ &   Dev$\downarrow$ & Test$\downarrow$ \\
        \midrule
        STMC~\cite{stmc}&21.1&20.7&19.6&21.0&-&-\\
        C$^2$SLR~\cite{zuo2022improving} & 20.5 & 20.4 & 20.2 & 20.4 & 31.9 & 31.0 \\
        SignBERT+~\cite{hu2023signbert+} & 19.9 & 20.0 & 18.8 & 19.9 & - & - \\
        SEN~\cite{hu2022self} & 19.5 & 21.0 & 19.3 & 20.7 & 31.1 & 30.7 \\
        CTCA~\cite{guo2023distilling} & 19.5 & 20.1 & 19.3 & 20.3 & 31.3 & 29.4 \\
        CorrNet~\cite{hu2023continuous} & 18.8 & 19.4 & 18.9 & 20.5 & 30.6 & 30.1 \\
        TwoStream~\cite{chentwo} & 18.4 & 18.8 & 17.7 & 19.3 & 25.4 & 25.3 \\

        \midrule
        Ours & \textbf{17.9} & \textbf{18.0} & \textbf{17.2} & \textbf{18.6} & \textbf{24.8} & \textbf{24.4} \\
        
    \bottomrule
    \end{tabular}
    }
    \vspace{-2mm}
    \caption{Comparison with other \textit{offline} CSLR methods.}
    \label{tab:sota_cslr}
    \vspace{-2mm}
\end{table}

\noindent\textbf{Offline Recognition.}
As described in Sec.~\ref{sec:ext} and illustrated in Fig.~\ref{fig:adapter}, our well-trained online model can enhance the performance of any offline model through the use of an adapter network. We instantiate the offline model with the TwoStream-SLR model due to its superior performance. As shown in Tab. \ref{tab:sota_cslr}, our approach, which involves fine-tuning only the lightweight adapter network and classification head, outperforms the previous best results by 0.8/0.7/0.9\% on the test sets of the three benchmarks.

\begin{table}[t]
\small
\setlength\tabcolsep{2pt}
    \centering
    \resizebox{\linewidth}{!}{
    \begin{tabular}{l|c|cccc}
    \toprule
    Method & Window Size & WER\%$\downarrow$ & AL$\downarrow$ & WPL$\downarrow$ & Memory$\downarrow$\\

    \midrule
    \multirow{5}{*}{TwoStream} & Entire video & 17.7 & 4,299 & 261 & 15.0 \\
    & 40 & 23.1 & 800 & 94 & 6.4 \\
    & 32 & 24.7 & 640 & 66 & 6.0 \\
    & 24 & 28.8 & 480 & 50 & 5.5 \\
    & 16 & 38.4 & 320 & 31 & 5.1 \\

    \midrule
    Ours & 16 & 22.2 & 320 & 29 & 5.1 \\
    
    \bottomrule
    \end{tabular}
    }
    \vspace{-3mm}
    \caption{Comparison with offline/online TwoStream-SLR in latency and memory cost (GB) on the P-2014T dev set. AL (ms): algorithmic latency; WPL (ms): window processing latency. It refers to offline recognition when the window size is set to ``Entire video.''}
    \vspace{-3mm}
    \label{tab:latency}
\end{table}

\vspace{-1mm}
\subsection{Ablation Studies}
\vspace{-1mm}
Unless otherwise specified, all ablation studies are conducted on P-2014T.

\noindent\textbf{Latency and Memory Cost.}
Offline models are hampered by high latency and substantial memory requirements. As Tab.~\ref{tab:latency} illustrates, we quantitatively compare our method against both offline and online TwoStream-SLR models concerning latency and memory costs. Following prior research in online speech recognition \cite{strimel2023lookahead, shi2021emformer}, we categorize latency into two types: algorithmic latency (AL) and window processing latency (WPL). AL refers to the minimum theoretical delay necessary for generating a prediction, which directly correlates with the window size. In contrast, WPL denotes the actual time required to produce a prediction for a specific window input. These evaluations are conducted using a single Nvidia V100 GPU. The findings highlight that, in comparison to the offline TwoStream model, our online model achieves a significant reduction in AL by approximately 92\% and lowers memory costs by 66\%. Additionally, our approach significantly surpasses the online TwoStream-SLR in performance using the same window size. A demo is available in the supplementary materials.

\begin{table}[t]
\centering
\small
\setlength\tabcolsep{2pt}
\resizebox{\linewidth}{!}{
\begin{tabular}{c|c|cc|c|cc}
\toprule
BG & Sign & \multicolumn{2}{c|}{Gloss-Level Training} & Sal. & \multirow{2}{*}{Dev$\downarrow$} & \multirow{2}{*}{Test$\downarrow$} \\
Class & Aug. & G-L Samp. & G-L Loss & Loss & & \\

\midrule
& & & & & 62.6 & 62.9 \\
\checkmark & & & & & 49.1 & 48.4 \\
\checkmark & \checkmark & & & & 24.4 & 24.4 \\

\checkmark & \checkmark & \checkmark & & & 22.7 & 23.4 \\
\checkmark & \checkmark & \checkmark & \checkmark & & 22.4 & 22.6 \\ 
\checkmark & \checkmark & \checkmark & \checkmark & \checkmark & \textbf{22.2} & \textbf{22.1} \\

\bottomrule
\end{tabular}
}
\vspace{-2mm}
\caption{Ablation studies for the major components. Each row employs the post-processing. BG: background; Aug.: augmentation; Samp.: sampling; Sal.: saliency.}
\label{tab:abl_main}
\vspace{-1mm}
\end{table}

\noindent\textbf{Effects of Major Components.}
In Tab. \ref{tab:abl_main}, we examine the effect of each major component by progressively adding them to our baseline ISLR model. This baseline model is trained only on pseudo GT ($\{s\}$) without the background class, using a single objective function $\mathcal{L}_{ce}^I$, achieving a WER of 62.9\% on the test set. Then, we introduce the background class into the training, resulting in a significant WER reduction of 14.5\%. The largest performance gain comes from sign augmentation: the model trained on both pseudo GT and augmented signs ($\{s\}\cup\{\hat{s}\}$) outperforms the model trained only on $\{s\}$, reducing the WER by 24.0\%. Next, our gloss-level training strategy, which uses: 1) a gloss-level sampling strategy that randomly selects $M$ glosses, each comprising $K$ instances; 2) an improved objective function $\mathcal{L}_{ce}^I + \mathcal{L}_{ce}^G$, further decreases the WER to 22.6\%. At last, adding the saliency loss will lead to the final test WER of 22.1\% with negligible extra costs.

\begin{table}[t]
\centering
\small
\begin{tabular}{l|ccc}
\toprule
Method & Accuracy$\uparrow$ & Dev$\downarrow$ & Test$\downarrow$ \\

\midrule
Equal Partitions & 14.4 & 92.6 & 92.3 \\
CTC Forced Alignment & \textbf{93.4} & \textbf{22.2} & \textbf{22.1} \\

\bottomrule
\end{tabular}
\vspace{-2mm}
\caption{Study on sign segmentor.}
\vspace{-3mm}
\label{tab:sign_seg}
\end{table}

\noindent\textbf{Sign Segmentor.}
We segment isolated signs from continuous videos to build a dictionary for ISLR model training. It is infeasible to directly evaluate its quality due to the lack of frame-level annotations. As an alternative, we invite an expert signer to conduct a human evaluation on the isolated signs of 100 randomly picked glosses in P-2014T. The signer needs to judge whether each sign is correctly categorized. As shown in Tab. \ref{tab:sign_seg}, our default sign segmentor that uses the CTC forced alignment algorithm can lead to an accuracy of 93.4\%. To better validate its significance, we implement a baseline sign segmentor that equally partitions each continuous sign video according to the number of glosses. The isolated signs obtained in this way suffer from low accuracy (14.4\%), resulting in a much worse online CSLR performance (>90\% WER) than our default strategy.

\begin{table}[t]
\small
    \centering
    \begin{tabular}{cc|cc}
    \toprule
    Strategy & Threshold & Dev$\downarrow$ & Test$\downarrow$ \\
    
    \midrule
    IoU & 0.5 & 27.4 & 27.1 \\
    IoU & 0.3 & 23.4 & 23.6 \\
    Center & N/A & \textbf{22.2} & \textbf{22.1} \\
    
    \bottomrule
    \end{tabular}
    \vspace{-2mm}
    \caption{Study on sign augmentation strategies.}
    \vspace{-1mm}
    \label{tab:iou}
\end{table}

\noindent\textbf{Sign Augmentation.}
As described in Sec.~\ref{sec:dict}, we augment each segmented isolated sign $s$ (pseudo ground truth) by generating a collection of video clips $\{\hat{s}\}$ around it. These clips are centered within the duration of $s$. We compare our default strategy with an alternative that employs an intersection-over-union (IoU) criterion~\cite{shou2016temporal} to generate augmented sign clips. In this IoU-based strategy, clips $\{\hat{s}\}$ are selected if they meet the condition IoU$(s, \hat{s})\geq\gamma$, where $\gamma$ is a predefined threshold. As shown in Tab.~\ref{tab:iou}, the IoU-based strategy is sensitive to the threshold variation: a large threshold may result in insufficient augmented signs, particularly for short isolated signs. In contrast, our default strategy does not rely on a predefined threshold and considers each isolated sign $s$ equally.

\begin{table}[t]
\centering
\small
    \begin{tabular}{l|ccccc}
    \toprule
    $\lambda$ & 1.0 & 0.7 & 0.5 & 0.3 & 0.0 \\
    
    \midrule
    Dev$\downarrow$ & 20.5 & 17.3 & \textbf{17.2} & 17.4 & 17.7 \\
    Test$\downarrow$ & 20.8 & 18.7 & \textbf{18.6} & \textbf{18.6} & 19.3 \\
    
    \bottomrule
    \end{tabular}
    \vspace{-2mm}
    \caption{Study on fusion weight $\lambda$.}
    \vspace{-2mm}
    \label{tab:fuse_weight}
\end{table}

\noindent\textbf{Feature Fusion.}
In Tab. \ref{tab:fuse_weight}, we study the fusion weight $\lambda$, when combining the features produced by our online model with an adapter network and the offline model (\ie, TwoStream-SLR~\cite{chentwo}) to boost offline recognition (see Sec.~\ref{sec:ext}). Setting $\lambda=0.0$ degenerates the integrated model to the offline TwoStream-SLR model, whereas $\lambda=1.0$ indicates that only the features encoded by the online model are used. Note that when $\lambda=1.0$, the fused model is trained using the original ISLR model (whose parameters are frozen), the adapter network, and the classification head, under the supervision of the CTC loss. Thus, the resulting model performs (20.5/20.8\% WER on dev/test set) better than its online counterpart (22.2/22.1\% WER on dev/test set), as it considers more contexts during training. We empirically set $\lambda=0.5$.
More ablation studies are available in Sec.~\ref{sec:more_exp}.

\vspace{-1mm}
\section{Conclusion}
\vspace{-1mm}
In this work, we develop a practical online CSLR framework. First, we construct a sign dictionary that aligns with the glossary of a target dataset. To enrich the training data, we collect augmented signs by cropping clips around each sign. To enable online CSLR, we train an ISLR model on the dictionary, using both standard classification loss and the introduced saliency loss. During inference, we perform online CSLR by feeding each sliding window into the well-optimized ISLR model on the fly. A simple yet efficient post-processing algorithm is introduced to eliminate duplicate predictions. Furthermore, two extensions are proposed for online SLT and boosting offline CSLR models, respectively. 
Along with the extensions, our framework achieves SOTA performance across three benchmarks under various task settings.

\section{Limitations}
Although we present an effective system for online CSLR, our approach has several limitations. 
First, we use a sign segmentor, \ie, a pre-trained CTC-based CSLR model, to segment continuous sign videos into several isolated sign clips as pseudo GT when building a dictionary. However, the boundaries of signs are inherently ambiguous, introducing unavoidable noise into the subsequent ISLR model training. Recent works on sign segmentation \cite{renz2021sign,moryossef2023linguistically} may assist in constructing a dictionary with high-quality samples, which are expected to benefit our system. 
Second, as discussed in TwoStream-SLR \cite{chentwo}, imprecise 2D keypoint estimation caused by motion blur, low-quality video, etc., can degrade model performance. A keypoint estimator specifically designed for sign language recognition might mitigate this issue.
Third, our training framework certainly causes losses of contextual information, resulting in worse performance than typical offline models. In the future, introducing extra knowledge, \eg, facial expressions \cite{viegas2023facial}, handshape \cite{zhang2023handshape}, and phonology \cite{kezar2023phonology}, may boost model performance.

It is also worth mentioning that our method relies on glosses. As suggested by \cite{muller2023considerations}, below we discuss the limitations of gloss-dependent approaches and the Phoenix datasets:
\begin{itemize}
\setlength{\itemsep}{0pt}
\setlength{\parsep}{0pt}
\setlength{\parskip}{1pt}
    \item Limitations of gloss-dependent approaches. Sign languages convey information through both manual and non-manual features. This multi-channel nature makes glosses—unique identifiers for signs—prone to irrecoverable information loss. Our work represents an initial effort in online sign language processing, and we hope it will inspire future research towards more advanced, gloss-free sign language translation.

    \item Limitations of the Phoenix datasets. We recognize several limitations in the Phoenix-2014 and Phoenix-2014T datasets, such as: 1) a narrow focus on weather forecasts; 2) a limited number of video-sentence pairs; and 3) simplistic glosses that lose non-manual features. To better assess our method's effectiveness, we introduce CSL-Daily, the largest Chinese sign language dataset, with a vocabulary of 2,000 glosses and about 20,000 video-sentence pairs—almost 2.5 times larger than Phoenix. CSL-Daily covers more diverse domains, such as family life and medical care. 
\end{itemize}

\bibliography{main}

\begin{algorithm*}[!ht]
\caption{Search for the optimal alignment path}\label{alg:segment} 
\begin{algorithmic}[1]
\State \textbf{Input:} Frame-wise probabilities $\boldsymbol{p}$; Extended gloss sequence $\hat{\boldsymbol{g}}$; Initialized $q(t,0)$ and $q(1,n)$
\State \textbf{Output:} The optimal alignment path $\boldsymbol{\theta}^*=(\theta^*_{1},\ldots,\theta^*_{T})$

\For{$n\gets 1 \ \textrm{to} \ 2N+1 $}
\Comment{Recursive computation}
    \For{$t\gets 2 \ \textrm{to} \ T $} 
        \State $q(t,n)=p_t(\hat{g}_n)\max_{f(n) \leq k \leq n} q(t-1,k)$        
    \EndFor
\EndFor
\State $n \gets \argmax_{k\in\{2N,2N+1\}}q(T,k)$ \Comment{Backtracking}
\State $\theta^*_{T} \gets  \hat{g}_n$
\For{$t\gets T-1 \ \textrm{to} \ 1$}
\State $n\gets \argmax_{f(n)\leq k \leq n}q(t,k)$
\State $\theta^*_{t} \gets \hat{g}_n$
\EndFor
\State \textbf{return} $\boldsymbol{\theta}^*=(\theta^*_{1},\ldots,\theta^*_{T})$\
\end{algorithmic}
\end{algorithm*}


\appendix

\section{More Implementation Details}
\subsection{Sign Segmentor}
\label{sec:sign_seg}

\begin{figure}[h]
\centering
\includegraphics[width=1.0\linewidth]{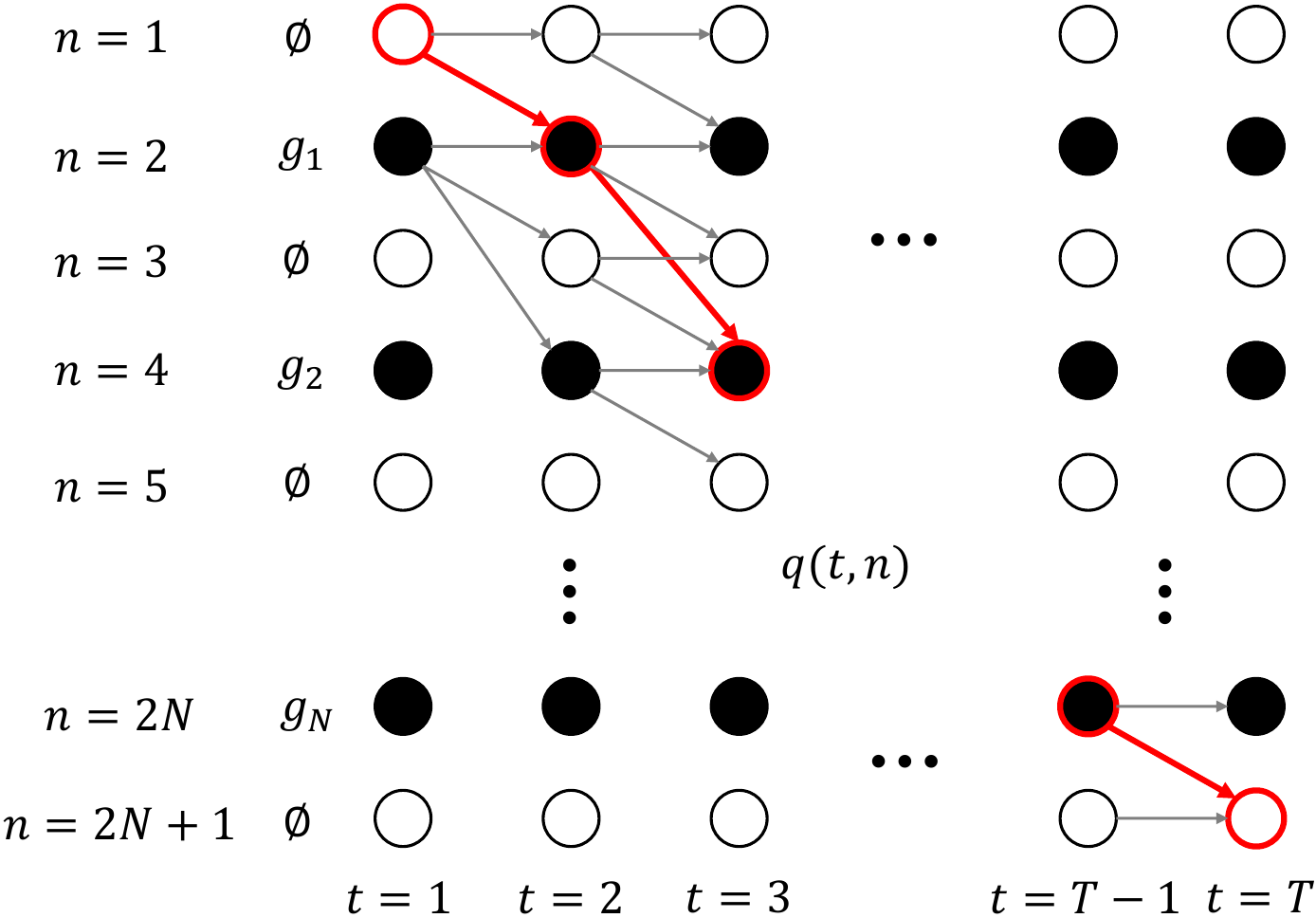}
\caption{Illustration of the CTC forced alignment algorithm used to compute $q(t,n)$ (Eq. \ref{eq:var}). $\varnothing$ is the blank class, $(g_1, \dots, g_n)$ is the gloss sequence. The red lines denote the optimal path, which is obtained by backtracking from the final gloss that has the maximum probability (Eq. \ref{eq:final_prob}). Pseudo code is available in Alg. \ref{alg:segment}.}
\label{fig:dtw}
\end{figure}

We use a pre-trained CSLR model, TwoStream-SLR \cite{chentwo}, to segment continuous sign language videos into a set of isolated sign clips, which are then utilized to train our ISLR model. Below we formulate the segmentation process.

Given a continuous sign video $\boldsymbol{V}$ comprising $T$ frames and its gloss sequence $\boldsymbol{g}=(g_1,\dots, g_{N})$ consisting of $N$ glosses, the probability of an alignment path $\boldsymbol{\theta} = (\theta_1, \dots, \theta_T)$ with respect to the ground truth $\boldsymbol{g}$, where $\theta_t \in \{g_i\}_{i=1}^N \cup \{\varnothing\}$ and $\varnothing$ is the blank (background) class, can be estimated with the conditional independence assumption:
\begin{equation}
\label{eq:cond_prob}
    p(\boldsymbol{\theta}|\boldsymbol{V}) = \prod_{i=1}^T p_t(\theta_t),
\end{equation}
where $p_t(\theta_t)$ denotes the posterior probability that the $t$-th frame is predicted as the class $\theta_t$. Note that due to the temporal pooling layers in the model's backbone (S3D \cite{xie2018rethinking}), we up-sample the original output probabilities of the CSLR model by a factor of 4 to match the length of the input sign video.

The optimal path is the one with the maximum probability: 
\begin{equation}
    \boldsymbol{\theta}^* = \argmax_{\boldsymbol{\theta} \in \mathcal{S}(\boldsymbol{g})} p(\boldsymbol{\theta}|\boldsymbol{V}),
\end{equation}
where $\mathcal{S}(\boldsymbol{g})$ denotes the set containing all feasible alignment paths with respect to ground truth $\boldsymbol{g}$. After obtaining the optimal path $\boldsymbol{\theta}^*$, we aggregate successive duplicate predictions into a single isolated sign.

We apply the CTC forced alignment algorithm \cite{dnf,ctc,Wei_2023_ICCV} to search for the optimal path $\boldsymbol{\theta}^*$. First, we insert blanks to the gloss sequence following the practice in \cite{dnf, ctc}. This process results in an extended gloss sequence of length $2N+1$: $\hat{\boldsymbol{g}} = (\varnothing, g_1, \varnothing, g_2, \dots, \varnothing, g_N, \varnothing)$. Subsequently, we define $q(t,n)$ as the maximum probability, up to time step $t$, for the sequence comprising the first $n$ elements of $\hat{\boldsymbol{g}}$. $q(t,n)$ can be recursively computed as:
\begin{equation}
\label{eq:var}
    q(t,n) = p_t(\hat{g}_n) \max_{f(n)\leq k \leq n} q(t-1,k),
\end{equation}
where
\begin{equation}
f(n)=
\begin{cases}
    \ n-1 \quad \text{if } \hat{g}_n=\varnothing \text{ or } \hat{g}_{n-2} = \hat{g}_n \\
    \ n-2 \quad \text{otherwise}
\end{cases}
\end{equation}
following \cite{ctc}. The initial conditions of $q(t,n)$ are defined as:
\begin{align}
&q(t,0) = 0, \ 1\leq t \leq T, \\
&q(1,n) =
\begin{cases}
    \ p_1(\hat{g}_n) \  & n=1,2 \\
    \ 0 & 2 < n \leq 2N+1
\end{cases}
.
\end{align}
The probability of the optimal path can be formulated as:
\begin{equation}
\label{eq:final_prob}
    p(\boldsymbol{\theta}^*|\boldsymbol{V}) = \max_{k\in\{2N,2N+1\}} q(T,k).
\end{equation}
Finally, the optimal path $\boldsymbol{\theta}^*$ can be obtained by backtracking $p(\boldsymbol{\theta}^*|\boldsymbol{V})$ in Eq. \ref{eq:final_prob}. We provide an illustration and pseudo code of both the recursive computation and backtracking in Fig. \ref{fig:dtw} and Alg.~\ref{alg:segment}.

\subsection{Online Inference}
\label{sec:inf_detail}

\begin{algorithm*}[t]
\caption{Post-processing for online inference}
\label{alg:post} 
\begin{algorithmic}[1]
\State \textbf{Input:} ISLR model $\mathcal{M}$; sliding window size $W$; sliding stride $S$; voting bag size $B$
\State \textbf{Output:} Post-processed predictions
\State $i \gets 0$
\State $raw \gets \mathrm{Queue}(maxsize=B)$ \Comment{Raw predictions}
\State $temp \gets \varnothing$  \Comment{Variable to store last voting result}
\State $output \gets [\varnothing]$ \Comment{Post-processed predictions}

\While{receive new frames}
\State $V \gets \mathrm{concat}(frame_{i},\dots,frame_{i+W-1})$ 
\State $p \gets \arg\max(\mathcal{M}(V))$
\State $raw.\mathrm{push}(p)$

\If{$raw.\mathrm{full}()$}
\State $p_v \gets \mathrm{voting}(raw)$ \Comment{Majority voting}
\label{code:voting}
\If{$(p_v \neq \varnothing) \text{ and } (p_v \neq output[-1] \text{ or } temp = \varnothing)$} \Comment{$[-1]$ denotes the last element}
\label{code:drop_blk}
\State $\mathrm{print}(p_v)$
\Comment{Output online predictions}
\State $output.\mathrm{append}(p_v)$
\EndIf
\State $temp \gets p_v$
\State $raw.\mathrm{pop}()$

\EndIf
\State $i \gets i+S$
\EndWhile

\State \textbf{return} $output[1\text{:}]$
\Comment{Output final predictions}
\end{algorithmic}
\end{algorithm*}

The pseudo code of our online post-processing algorithm is provided in Alg. \ref{alg:post}. The algorithm has two key functions: (1) voting-based deduplication (Line \ref{code:voting}), and (2) background elimination (Line \ref{code:drop_blk}). 
We build a simple deduplicator based on majority voting: we collect predictions from $B$ sliding windows to form a voting bag, and output the predicted class that appears more than $B/2$ times. If no class meets this criterion, the bag yields a background class $\varnothing$. Finally, we eliminate background predictions and merge non-background predictions; for instance, $\{A,\varnothing,\varnothing\}\rightarrow\{A\}$ and $\{A,A,A\}\rightarrow\{A\}$.

\subsection{Architecture of the ISLR Model}
Following TwoStream-SLR \cite{chentwo}, we build our ISLR model using a two-stream architecture, which processes both RGB videos and keypoint heatmaps to more effectively interpret sign languages. The video stream consists of a S3D network \cite{xie2018rethinking} for feature extraction, coupled with a head network. The head network includes a temporal average pooling layer and a fully connected layer followed by a softmax layer for computing gloss probabilities. The input video dimensions are $T\times H\times W\times 3$, where $T$ represents the number of frames, and $H$ and $W$ denote the frame height and width, respectively. We standardize $H$ and $W$ to 224 and set $T$ to 16. The S3D network outputs features of size $T/8\times 1024$ after spatial pooling, which are then input into the head network to generate the gloss probabilities.

Human keypoints are represented as a sequence of heatmaps, following \cite{duan2022revisiting}, allowing the keypoint stream to share the same architecture as the video stream. For each sign video, we first use HRNet \cite{sun2019deep} pre-trained on COCO-WholeBody \cite{jin2020whole}, to generate 11 upper body keypoints, 42 hand keypoints, and 10 mouth keypoints. These extracted keypoints are then converted into heatmaps using a Gaussian function \cite{chentwo, duan2022revisiting}. The input heatmap sequence has dimensions $T\times H'\times W' \times N_k$, where $N_k=63$ denotes the total number of keypoints, and we set $H'=W'=112$ to reduce computational cost.

Following TwoStream-SLR \cite{chentwo}, we incorporate bidirectional lateral connections \cite{duan2022revisiting, feichtenhofer2019slowfast} and a joint head network to better explore the potential of the two-stream architecture. Lateral connections are applied to the output features of the first four blocks of S3D. Specifically, we utilize strided convolution and transposed convolution layers with a kernel size of $3\times 3$ to align the spatial resolutions of features produced by the two streams. Subsequently, we add the mapped features from one stream to the raw output features of the other stream to achieve information fusion. The joint head network maintains the architecture of the original network in each stream. Its distinctive feature is that it processes the concatenation of the output features of both streams. Refer to the original TwoStream-SLR paper \cite{chentwo} for additional details.

\subsection{Training Details}
\noindent\textbf{ISLR Model.}
We train our ISLR model for 100 epochs with an effective batch size of $4\times6$, which means that 4 glosses and 6 instances for each gloss are sampled. For data augmentation, we use spatial cropping with a range of [0.7-1.0] and temporal cropping. Both RGB videos and heatmap sequences undergo identical augmentations to maintain spatial and temporal consistency. We employ a cosine annealing schedule and an Adam optimizer \cite{adam} with a weight decay of $1e-3$ and an initial learning rate of $6e-4$. Label smoothing is applied with a factor of 0.2. All models are trained on 8$\times$ Nvidia V100 GPUs.

\noindent\textbf{Wait-$\boldsymbol{k}$ Gloss-to-Text Network.}
To facilitate online sign language translation, we train an additional gloss-to-text network using the wait-$k$ policy \cite{ma2019stacl}, setting $k=2$ as suggested in \cite{yin2021simulslt}. We employ the mBART architecture \cite{mbart} for this network, owing to its proven effectiveness in gloss-to-text translation \cite{chentwo}. The implementation of the wait-$k$ policy strictly adheres to the guidelines in \cite{ma2019stacl}, involving the application of causal masking. The network undergoes training for 80 epochs, starting with an initial learning rate of $1e-5$. To prevent overfitting, we incorporate dropout with a rate of 0.3 and use label smoothing with a factor of 0.2.

\noindent\textbf{Boosting Offline Model.}
Our online model can boost the performance of offline models with an adapter, as shown in Fig. 4 of the main paper. When fine-tuning the adapter network and the classification head, we adopt a smaller learning rate of $1e-4$ and fewer epochs of 40, and the weight $\lambda=0.5$. We adopt the CTC loss \cite{ctc} as our objective function. Eq. \ref{eq:cond_prob} computes the probability of a single alignment path $\boldsymbol{\theta}$. The CTC loss is applied across the set of all feasible alignment paths $\mathcal{S}(\boldsymbol{g})$ in relation to the ground truth $\boldsymbol{g}$:
\begin{equation}
    \mathcal{L}_{ctc} = -\log \sum_{\boldsymbol{\theta} \in \mathcal{S}(\boldsymbol{g})} p(\boldsymbol{\theta}|\boldsymbol{V}).
\end{equation}

\begin{table}[t]
\small
    \centering
    \begin{tabular}{l|cc|cc}
    \toprule
    \multirow{2}{*}{Method} & \multicolumn{2}{c|}{P-2014} & \multicolumn{2}{c}{P-2014T} \\
    & Dev$\downarrow$ & Test$\downarrow$ & Dev$\downarrow$ & Test$\downarrow$ \\

    \midrule
    NLA-SLR & 34.2 & 33.7 & 32.9 & 33.4  \\
    Ours & \textbf{22.6} & \textbf{22.1} & \textbf{22.2} & \textbf{22.1} \\

    \bottomrule
    \end{tabular}
    \caption{Comparison with the state-of-the-art ISLR method, NLA-SLR \cite{zuo2023natural}, in the online CSLR context.}
    \label{tab:nla-slr}
\end{table}

\section{More Quantitative Results}
\label{sec:more_exp}
\subsection{Comparison with the SOTA ISLR Method}
In this paper, we utilize an ISLR model to fulfill online CSLR. Our approach introduces novel techniques for enhancing the training of the ISLR model, including sign augmentation, gloss-level training, and saliency loss. To further verify the effectiveness of our model, we re-implement the leading ISLR method, NLA-SLR \cite{zuo2023natural}. This approach integrates natural language priors into ISLR model training and demonstrates state-of-the-art results across various ISLR benchmarks \cite{li2020word, joze2019ms, hu2021global}. However, as shown in Tab.~\ref{tab:nla-slr}, in the online CSLR context, our method significantly outperforms NLA-SLR, evidenced by a notable 11.3\% reduction in word error rate (WER) on the Phoenix-2014T test set, affirming the superiority of our ISLR techniques.

\subsection{Gloss-to-Text Translation Using GT Glosses}
\begin{table}[t]
\small
    \centering
    \begin{tabular}{l|c|cc}
    \toprule
        Method & Win. Size & Dev$\uparrow$ & Test$\uparrow$ \\
    \midrule
        GT Glosses & N/A & 25.41 & 24.49  \\
    \midrule
        \multirow{4}{*}{TwoStream} & 40 & 22.80 & 22.64 \\
        & 32 & 22.23 & 22.01 \\
        & 24 & 22.19 & 19.92 \\
        & 16 & 18.36 & 18.81 \\
    \midrule
        Ours & 16 & \textbf{23.75} & \textbf{23.69} \\
    \bottomrule
    \end{tabular}
    \vspace{-2mm}
    \caption{Comparison with a gloss-to-text translation model using ground-truth glosses on P-2014T.}
    \label{tab:gt_gls}
    \vspace{-3mm}
\end{table}

As indicated in Tab. \ref{tab:gt_gls}, utilizing ground-truth glosses achieves a BLEU-4 score of 24.49 on the P-2014T test set. Notably, our online method approaches this upper bound more closely than other online TwoStream baselines, with a minimal gap of 0.8 BLEU-4 point.

\subsection{Study on Hyper-Parameters}
In Tab. \ref{tab:bg_ratio}, we vary the percentage of background samples used from 0\% to 100\%. We find that using all background samples yields the best performance. This result implies the effectiveness of incorporating the background class in modeling co-articulations. 

In a mini-batch, we randomly sample $M$ glosses, with each gloss comprising $K$ instances. The impact of varying $M$ and $K$ is explored in Tab. \ref{tab:num_gloss}. 

Our saliency loss aims to enforce the model to focus more on the foreground part. As detailed in Sec. 3.2 of the main paper, we upsample the feature by a factor of $\beta$. We examine various values of $\beta$ in Tab. \ref{tab:upsample}. 

We also investigate the optimal size of the sliding window in our proposed online CSLR method. Tab. \ref{tab:win_size} indicates that a window size of 16 frames is most effective, aligning closely with the average sign duration.

For our online post-processing, we implement majority voting to eliminate duplicates. The influence of the voting bag size $B$ is analyzed in Tab. \ref{tab:voting_size}. Here, $B=1$ implies the absence of post-processing. A moderate bag size is preferred as a larger bag might mistakenly drop correct predictions, leading to lower recall. Conversely, a smaller bag might not completely remove duplicates, resulting in lower precision.

\begin{table}[t]
\setlength\tabcolsep{4pt}
    \centering
    \begin{subtable}[t]{0.45\linewidth}
    \small
    \centering
    \begin{tabular}{c|cc}
    \toprule
    Perc. (\%) & Dev$\downarrow$ & Test$\downarrow$ \\
    
    \midrule
    0 & 45.8 & 46.8 \\
    20 & 25.4 & 25.1 \\
    50 & 22.5 & 23.7 \\
    100 & \textbf{22.2} & \textbf{22.1} \\
    
    \bottomrule
    \end{tabular}
    \caption{Percentage of background samples.}
    \label{tab:bg_ratio}
    \end{subtable}
    \hfill
    \begin{subtable}[t]{0.45\linewidth}
    \small
    \centering
    \begin{tabular}{cc|cc}
    \toprule
    $M$ & $K$ & Dev$\downarrow$ & Test$\downarrow$ \\
    
    \midrule
    12 & 2 & 22.7 & 24.0 \\
    6 & 4 & 22.5 & 22.8 \\
    4 & 6 & \textbf{22.2} & \textbf{22.1} \\
    
    \bottomrule
    \end{tabular}
    \caption{Number of glosses ($M$) and instances per gloss ($K$) in a mini-batch.}
    \label{tab:num_gloss}
    \end{subtable}
    
    \begin{subtable}[t]{0.45\linewidth}
    \small
    \centering
    \begin{tabular}{c|cc}
    \toprule
    $\beta$ & Dev$\downarrow$ & Test$\downarrow$ \\
    
    \midrule
    2 & 22.3 & 23.2 \\
    4 & \textbf{22.2} & \textbf{22.1} \\
    8 & \textbf{22.2} & 22.7\\
    
    \bottomrule
    \end{tabular}
    \caption{Up-sampling factor ($\beta$) in the saliency loss.}
    \label{tab:upsample}
    \end{subtable}
    \hfill
    \begin{subtable}[t]{0.45\linewidth}
    \small
    \centering
    \begin{tabular}{l|ccc}
    \toprule
    $W$ & Dev$\downarrow$ & Test$\downarrow$ \\
    
    \midrule
    8 & 25.9 & 25.6 \\
    16 & \textbf{22.2} & \textbf{22.1} \\
    32 & 23.0 & 23.2 \\
    
    \bottomrule
    \end{tabular}
    \caption{Sliding window size ($W$).}
    \label{tab:win_size}
    \end{subtable}
    
    \begin{subtable}[t]{0.9\linewidth}
    \small
    \centering
    \begin{tabular}{l|ccccccc}
    \toprule
    $B$ & 1 & 3 & 5 & 7 & 9 & 11 & 13 \\
    
    \midrule
    Dev$\downarrow$ & 54.8 & 27.8 & 23.0 & \textbf{22.2} & 23.4 & 26.1 & 31.7 \\
    Test$\downarrow$ & 57.3 & 29.0 & 23.2 & \textbf{22.1} & 23.2 & 25.9 & 31.5 \\
    
    \bottomrule
    \end{tabular}
    \caption{Voting bag size ($B$).}
    \label{tab:voting_size}
    \end{subtable}
    \caption{Studies on hyper-parameters.}
    \vspace{-3mm}
    \label{tab:abl_misc}
\end{table}

\begin{figure*}[h]
\begin{subfigure}[t]{0.47\textwidth}
\centering
\includegraphics[width=\textwidth]{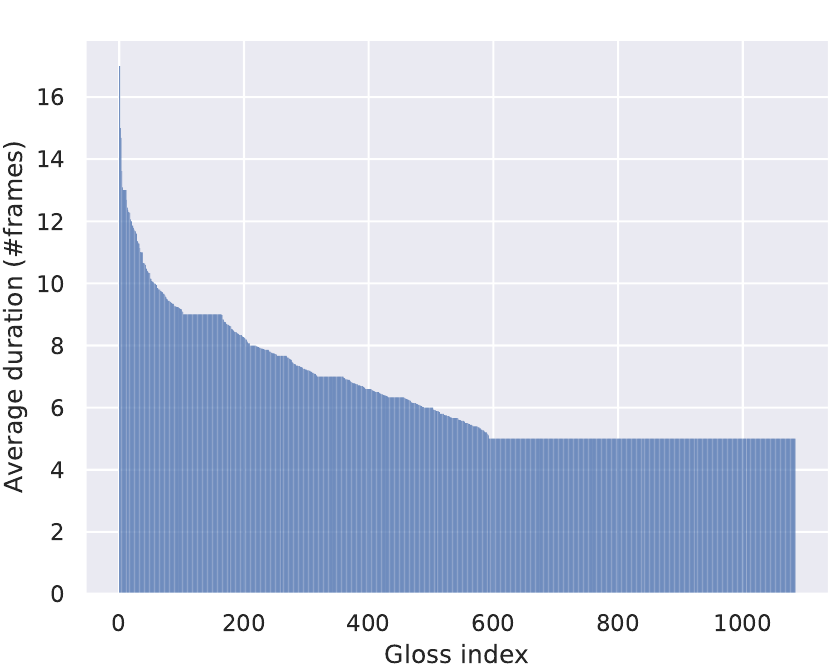}
\caption{}
\label{fig:a}
\end{subfigure}
\hfill
\begin{subfigure}[t]{0.47\textwidth}
\centering
\includegraphics[width=\textwidth]{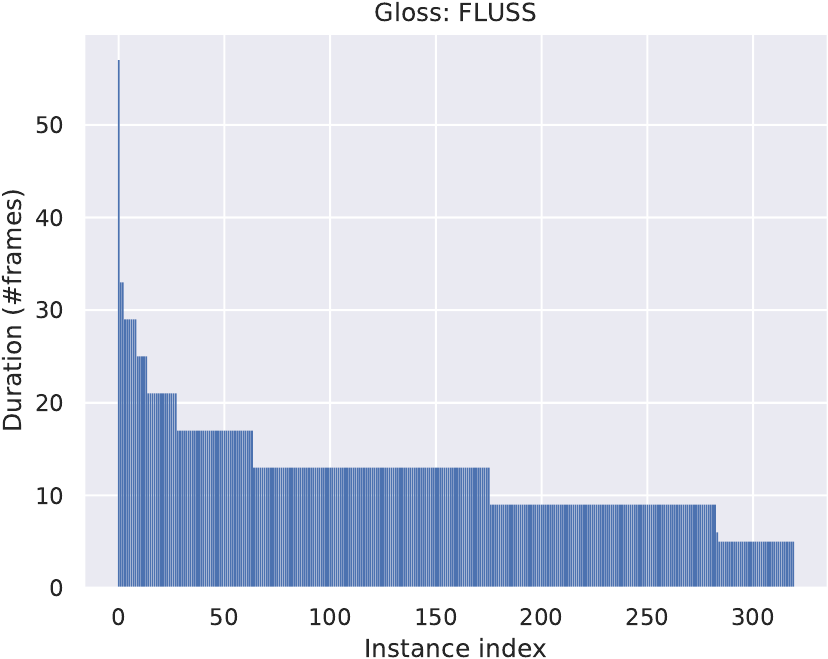}
\caption{}
\label{fig:b}

\end{subfigure}
\begin{subfigure}[t]{0.47\textwidth}
\centering
\includegraphics[width=\textwidth]{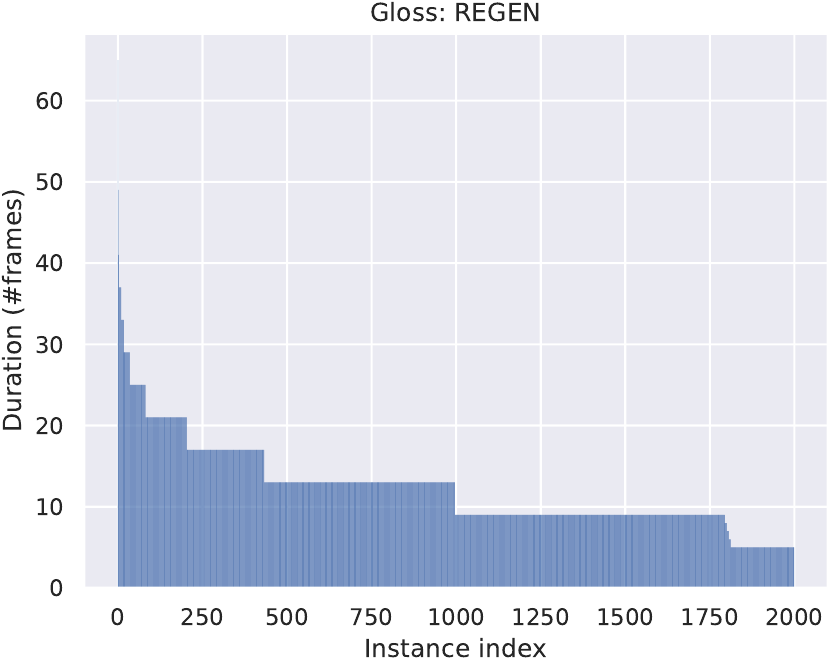}
\caption{}
\label{fig:c}
\end{subfigure}
\hfill
\begin{subfigure}[t]{0.47\textwidth}
\centering
\includegraphics[width=\textwidth]{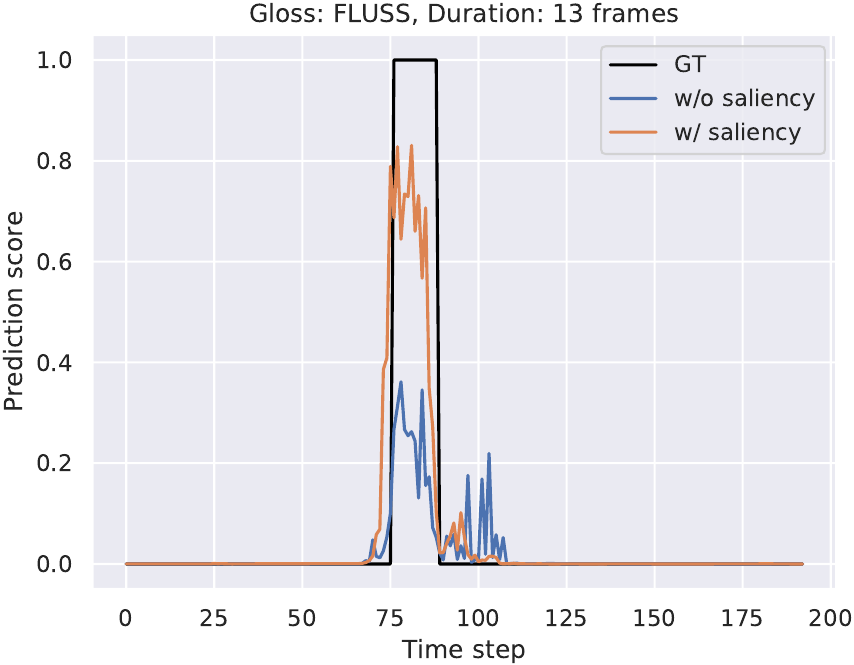}
\caption{}
\label{fig:d}
\end{subfigure}

\begin{subfigure}[t]{0.47\textwidth}
\centering
\includegraphics[width=\textwidth]{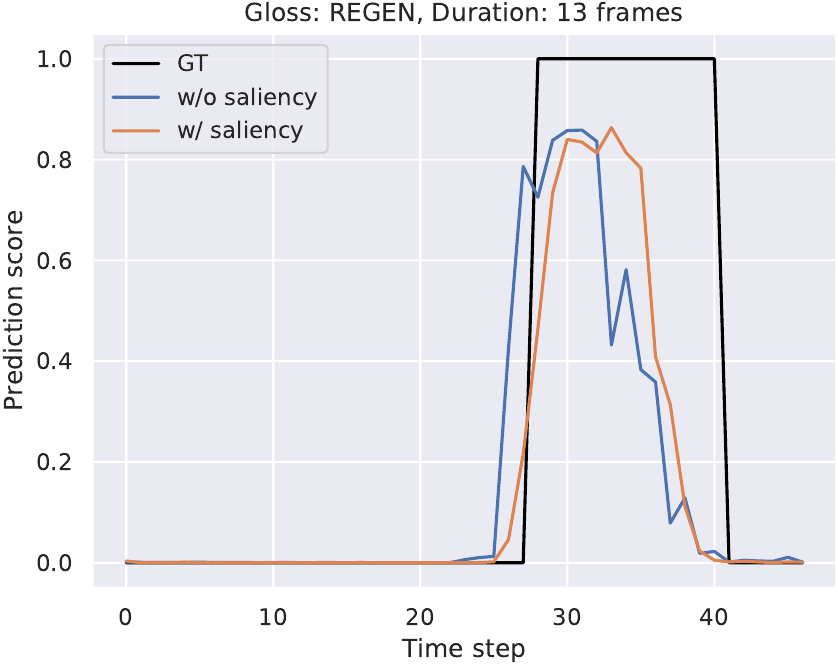}
\caption{}
\label{fig:e}
\end{subfigure}
\hfill
\begin{subfigure}[t]{0.47\textwidth}
\centering
\includegraphics[width=\textwidth]{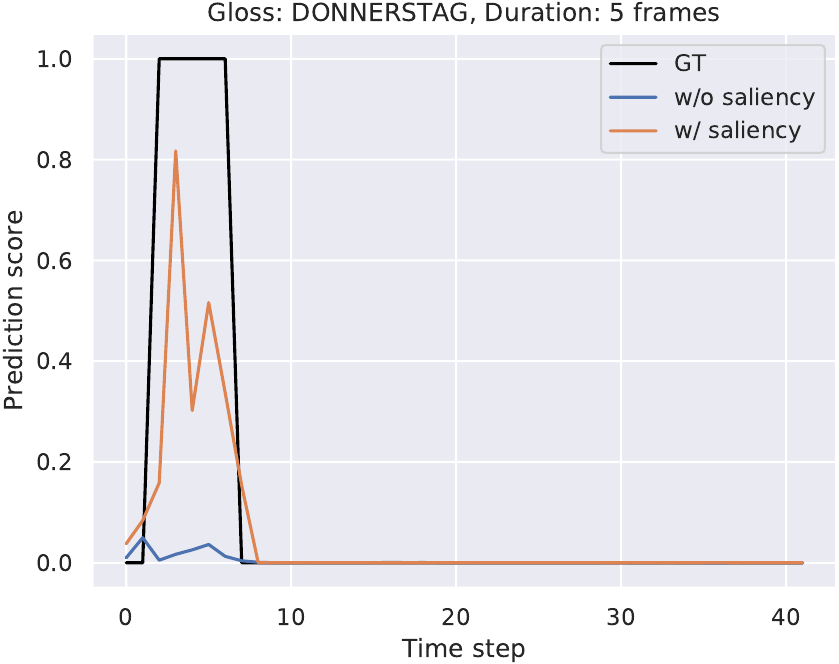}
\caption{}
\label{fig:f}
\end{subfigure}
\caption{Visualization of sign duration and prediction scores (output probabilities) on the Phoenix-2014T dev set. (a) Statistics of the average duration of each gloss in the vocabulary. To calculate the average duration of a specific gloss, we average the duration of all instances belonging to this gloss. (b)(c) Statistics of the sign duration at the instance level for two randomly selected glosses, namely ``FLUSS'' and ``REGEN''. (d)(e)(f) Window-wise prediction scores of three instances, each belonging to the glosses of ``FLUSS'', ``REGEN'' and ``DONNERSTAG'', respectively. Each time step is associated with a window center. We visualize the pseudo ground truths and the predictions made by the models with and without the use of saliency loss.}
\label{fig:sal_vis}
\end{figure*}

\section{Qualitative Results}
\subsection{Saliency Loss}
In general, a continuous sign video comprises multiple isolated signs linked together with meaningless transitional movements (co-articulations), each serving as a bridge between two adjacent signs. During inference, a given sliding window might include only a portion of an isolated sign, along with segments of one or two co-articulations. The variation in sign duration may also complicate this issue (Fig. \ref{fig:sal_vis}(a)(b)(c)). To enhance the model's ability to focus on the foreground signs, we introduce the saliency loss. Its objectives are to: 1) drive the model to assign higher activation to each foreground part; 2) encourage the model to learn more discriminative features of the foreground parts. In addition to demonstrating the improvement achieved by integrating the saliency loss, as shown in Tab. 6 of the main paper, we provide visualization results in Fig.~\ref{fig:sal_vis}(d)(e)(f). It is evident that, with the aid of the saliency loss, our model identifies foreground signs more precisely and yields higher activations when the sliding window encounters these signs.

\subsection{Comparison of Predictions}
As shown in Tab. \ref{tab:qual_res}, we conduct qualitative comparison between the online TwoStream-SLR and our approach, presenting three examples from the dev sets of Phoenix-2014T and CSL-Daily, respectively. It is clear that our proposed online model yields more accurate predictions than the online TwoStream-SLR, even when the latter uses a large window size of 40 frames.

\section{Broader Impacts}
Sign languages serve as the primary means of communication within the deaf community. Research on CSLR aims to bridge the communication gap between deaf and hearing individuals. While most existing CSLR research has concentrated on enhancing offline recognition performance, the development of an online framework remains largely unexplored. In this paper, we introduce a practical online solution that involves sequentially processing short video clips extracted from a sign stream. This is achieved by feeding these clips into a well optimized model for ISLR, thereby enabling online recognition. Our work, therefore, lays the groundwork for future advancements in online and real-time sign language recognition systems.

\begin{CJK*}{UTF8}{gbsn}
\begin{table*}[h]
\centering
\resizebox{0.99\linewidth}{!}{
\begin{tabular}{l|l|c}
\toprule
\textbf{Example (a)} & & WER\%$\downarrow$ \\
\midrule
\multirow{2}{*}{Ground truth} & TAG SUED MITTE WOLKE KRAEFTIG NEBEL & \multirow{2}{*}{-}\\
& (Day South Mid Cloud Heavy Fog) & \\
\midrule
Prediction ($W=40$) & TAG SUED MITTE \del{*****} \sub{MEISTENS} NEBEL & \multirow{2}{*}{33.3} \\
(TwoStream-SLR \cite{chentwo}) & (Day South Mid \del{*****} \sub{Mostly} Fog) & \\
\midrule
Prediction ($W=16$) & TAG SUED \ins{NEBEL} MITTE \sub{NEBEL} \sub{PUEBERWIEGEND} NEBEL & \multirow{2}{*}{50.0} \\
(TwoStream-SLR \cite{chentwo}) & (Day South \ins{Fog} Mid \sub{Fog} \sub{Overwhelmingly} Fog) & \\
\midrule
Prediction ($W=16$) & TAG SUED MITTE WOLKE KRAEFTIG NEBEL & \multirow{2}{*}{0.0} \\
(Ours)& (Day South Mid Cloud Heavy Fog) & \\
\midrule

\textbf{Example (b)} & & WER\%$\downarrow$ \\
\midrule
\multirow{2}{*}{Ground truth} & JETZT WETTER WIE-AUSSEHEN MORGEN DIENSTAG NEUNTE FEBRUAR & \multirow{2}{*}{-}\\
& (Now Weather Look Tomorrow Tuesday Ninth February) & \\
\midrule
Prediction ($W=40$) & JETZT WETTER WIE-AUSSEHEN MORGEN DIENSTAG \ins{WENN} NEUNTE FEBRUAR & \multirow{2}{*}{14.3} \\
(TwoStream-SLR \cite{chentwo})& (Now Weather Look Tomorrow Tuesday \ins{If} Ninth February) & \\
\midrule
Prediction ($W=16$) & JETZT WETTER \ins{JETZT} WIE-AUSSEHEN MORGEN DIENSTAG \del{******} \sub{FREUNDLICH} & \multirow{2}{*}{42.9} \\
(TwoStream-SLR \cite{chentwo})& (Now Weather \ins{Now} Look Tomorrow Tuesday \del{******} \sub{Friendly}) & \\
\midrule
Prediction ($W=16$) & JETZT WETTER WIE-AUSSEHEN MORGEN DIENSTAG NEUNTE FEBRUAR & \multirow{2}{*}{0.0} \\
(Ours)& (Now Weather Look Tomorrow Tuesday Ninth February) & \\
\midrule

\textbf{Example (c)} & & WER\%$\downarrow$ \\
\midrule
\multirow{2}{*}{Ground truth} & OST SUEDOST UEBERWIEGEND WOLKE BISSCHEN SCHNEE & \multirow{2}{*}{-}\\
& (East Southeast Mainly Cloud Bit Snow) & \\
\midrule
Prediction ($W=40$) & OST \del{*******} \sub{MEISTENS} WOLKE BISSCHEN SCHNEE & \multirow{2}{*}{33.3} \\
(TwoStream-SLR \cite{chentwo})& (East \del{*******} \sub{Mostly} Cloud Bit Snow) & \\
\midrule
Prediction ($W=16$) & \ins{REGION} \ins{KOMMEN} OST SUEDOST \sub{MEISTENS} WOLKE BISSCHEN SCHNEE & \multirow{2}{*}{50.0} \\
(TwoStream-SLR \cite{chentwo})& (\ins{Region} \ins{Come} East Southeast \sub{Mostly} Cloud Bit Snow) & \\
\midrule
Prediction ($W=16$) & OST SUEDOST \sub{MEISTENS} WOLKE BISSCHEN SCHNEE & \multirow{2}{*}{16.7} \\
(Ours)& (East Southeast \sub{Mostly} Cloud Bit Snow) & \\
\midrule

\textbf{Example (d)} & & WER\%$\downarrow$ \\
\midrule
\multirow{2}{*}{Ground truth} & 想\ 要\ 健康\ 吸烟\ 不 & \multirow{2}{*}{-}\\
& (Want Be Healthy Smoke No) & \\
\midrule
Prediction ($W=40$) & 想\ 要\ \ins{身体}\ 健康\ \ins{强}\ 吸烟\ 不 & \multirow{2}{*}{40.0} \\
(TwoStream-SLR \cite{chentwo})& (Want Be \ins{Body} Healthy \ins{Strong} Smoke No) & \\
\midrule
Prediction ($W=16$) & 想\ \ins{我}\ \sub{身体}\ 健康\ \ins{强}\ 吸烟\ \ins{你}\ 不 & \multirow{2}{*}{80.0} \\
(TwoStream-SLR \cite{chentwo})& (Want \ins{Me} \sub{Body} Healthy \ins{Strong} Smoke \ins{You} No) & \\
\midrule
Prediction ($W=16$) & 想\ 要\ \ins{身体}\ 健康\ 吸烟\ 不 & \multirow{2}{*}{20.0} \\
(Ours)& (Want Be \ins{Body} Healthy Smoke No) & \\
\midrule

\textbf{Example (e)} & & WER\%$\downarrow$ \\
\midrule
\multirow{2}{*}{Ground truth} & 今天\ 阴\ 大概\ 会\ 下雨 & \multirow{2}{*}{-}\\
& (Today Cloudy Probably Will Rain) & \\
\midrule
Prediction ($W=40$) & 今天\ \ins{来}\ 阴\ 大概\ 会\ 下雨 & \multirow{2}{*}{20.0} \\
(TwoStream-SLR \cite{chentwo})& (Today \ins{Come} Cloudy Probably Will Rain) & \\
\midrule
Prediction ($W=16$) & 今天\ \ins{来}\ 阴\ \ins{你}\ 大概\ 会\ 下雨 & \multirow{2}{*}{40.0} \\
(TwoStream-SLR \cite{chentwo})& (Today \ins{Come} Cloudy \ins{You} Probably Will Rain) & \\
\midrule
Prediction ($W=16$) & 今天\ 阴\ 大概\ 会\ 下雨 & \multirow{2}{*}{0.0} \\
(Ours)& (Today Cloudy Probably Will Rain) & \\
\midrule

\textbf{Example (f)} & & WER\%$\downarrow$ \\
\midrule
\multirow{2}{*}{Ground truth} & 明天\ 考试\ 要\ 带\ 笔\ 不\ 带\ 手机 & \multirow{2}{*}{-}\\
& (Tomorrow Exam Need Bring Pen No Bring Cellphone) & \\
\midrule
Prediction ($W=40$) & 明天\ \ins{买}\ 考试\ 要\ 带\ \ins{你}\ \sub{作业}\ 不\ 带\ 手机 & \multirow{2}{*}{37.5} \\
(TwoStream-SLR \cite{chentwo})& (Tomorrow \ins{Buy} Exam Need Bring \ins{You} \sub{Homework} No Bring Cellphone) & \\
\midrule
Prediction ($W=16$) & 明天\ \ins{买}\ 考试\ \ins{我}\ 要\ 带\ \ins{什么}\ \ins{快}\ \sub{作业}\ 不\ 带\ 手机 & \multirow{2}{*}{62.5} \\
(TwoStream-SLR \cite{chentwo})& (Tomorrow \ins{Buy} Exam \ins{Me} Need Bring \ins{What} \ins{Quickly} \sub{Homework} No Bring Cellphone) & \\
\midrule
Prediction ($W=16$) & 明天\ 考试\ 要\ 带\ \sub{作业}\ 不\ 带\ 手机 & \multirow{2}{*}{12.5} \\
(Ours)& (Tomorrow Exam Need Bring \sub{Homework} No Bring Cellphone) & \\

\bottomrule
\end{tabular}
}
\caption{Qualitative comparison between the online TwoStream-SLR and our approach on the dev sets of Phoenix-2014T (Example (a,b,c)) and CSL-Daily (Example (d,e,f)), respectively, under the online scenario. We use different colors to represent \sub{substitutions}, \del{deletions}, and \ins{insertions}, respectively. $W$ denotes the sliding window size.}
\label{tab:qual_res}
\end{table*}
\end{CJK*}

\end{document}